\documentclass[runningheads]{llncs}
\usepackage{graphicx}

\usepackage{tikz}
\usepackage{comment}
\usepackage{amsmath,amssymb} 
\usepackage{color}

\usepackage[accsupp]{axessibility}  

\usepackage{booktabs}
\usepackage{subcaption}
\usepackage{multirow}
\usepackage{enumitem}
\usepackage{array}
\usepackage[pagebackref,breaklinks,colorlinks]{hyperref}

\usepackage{epsfig}
\usepackage{graphicx}
\usepackage{amsmath}
\usepackage{amssymb}
\usepackage{bm}
\usepackage{threeparttable}
\usepackage{soul}
\usepackage{wrapfig}
\usepackage{sidecap}

\newcommand{\etal}{\textit{et al.~}}
\newcommand{\eg}{\textit{e.g.,~}}
\newcommand{\ie}{\textit{i.e.,~}}

\begin{document}
\pagestyle{headings}
\mainmatter
\def\ECCVSubNumber{1517}  

\title{Multi-Query Video Retrieval} 

\titlerunning{ }
%
\author{Zeyu Wang \and
Yu Wu \and
Karthik Narasimhan \and 
Olga Russakovsky}
\authorrunning{ }
%
\institute{Princeton University \\
\email{\{zeyuwang,yuwu,karthikn,olgarus\}@cs.princeton.edu}\\
}
\maketitle

\begin{abstract}

    Retrieving target videos based on text descriptions is a task of great practical value and has received increasing attention over the past few years. Despite recent progress, imperfect annotations in existing video retrieval datasets have posed significant challenges on model evaluation and development. In this paper, we tackle this issue by focusing on the less-studied setting of multi-query video retrieval, where multiple descriptions are provided to the model for searching over the video archive. We first show that multi-query retrieval task effectively mitigates the dataset noise introduced by imperfect annotations and better correlates with human judgement on evaluating retrieval abilities of current models. We then investigate several methods which leverage multiple queries at training time, and demonstrate that the multi-query inspired training can lead to superior performance and better generalization. We hope further investigation in this direction can bring new insights on building systems that perform better in real-world video retrieval applications.\footnote{Code is available at \url{https://github.com/princetonvisualai/MQVR}.}

\end{abstract}

\section{Introduction}

    With the vast amount of videos available online and new ones being generated and uploaded everyday, an efficient video search engine  is of great practical value. The core of a video search engine is the retrieval task, which deals with finding the best matching videos based on the user's text query~\cite{mithun2018learning,miech2018learning,liu2019use,dzabraev2021mdmmt,cheng2021improving}. With collective efforts over the years, current state-of-the-art methods~\cite{andres2021straightforward,luo2021clip4clip,fang2021clip2video} have achieved reasonable performance on several different video retrieval benchmarks~\cite{xu2016msr,chen2011collecting,anne2017localizing,krishna2017dense,rohrbach2015dataset}.

    \begin{figure}[t]
      \centering
      \includegraphics[width=0.9\linewidth]{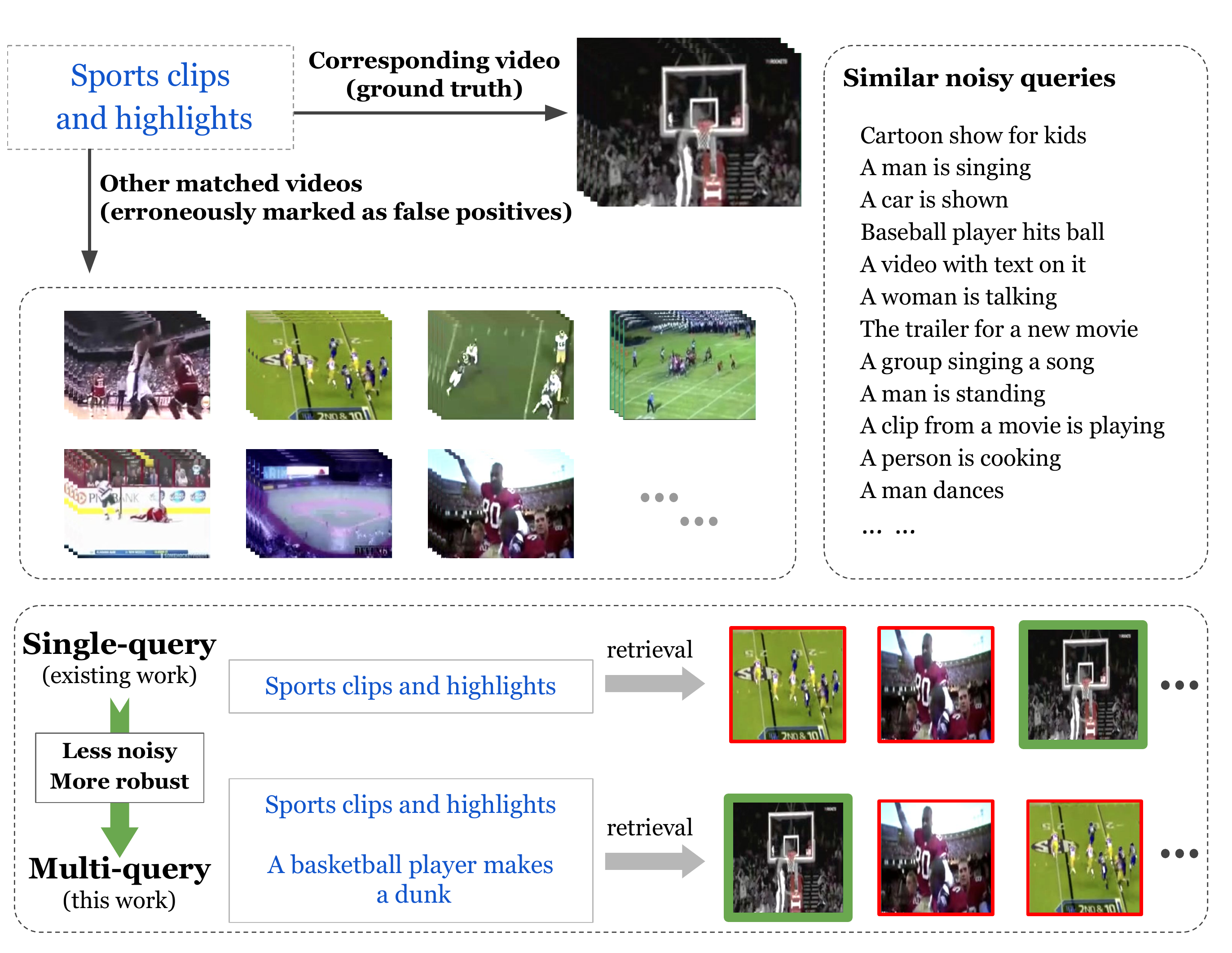}
      \caption{Existing video retrieval works focus on the single-query setting, where a single text description is input for retrieving target video. However, due to imperfect annotations in the datasets (too general descriptions that match many different videos, shown on top), retrieval model may be unfairly penalized for not finding the single target video. To counteract this under-specification issue, in this work we tackle the previously less-studied {\em multi-query} setting (bottom), which provides a less noisy benchmarking of retrieval abilities of current models.}
      \label{fig:pull}
    \end{figure}

    Despite these advances, existing works focus mostly on experimenting under the single-query setting, \ie retrieving target videos given a \textit{single} text description as input. A major issue with this paradigm is the problem of under-specification. As most existing video retrieval benchmarks are based on datasets collected for other tasks (\eg video captioning)~\cite{xu2016msr,wang2019vatex,chen2011collecting,krishna2017dense,rohrbach2015dataset}, lots of text descriptions are of low-quality and unsuited for video retrieval. Figure~\ref{fig:pull} top shows some examples of noisy annotations from one of the most widely-used video retrieval datasets MSR-VTT~\cite{xu2016msr}. One can notice that these annotations are very general and can be perfectly matched to lots of videos. This can cause trouble during evaluation as all matched videos would be given similar matching scores and essentially ranked randomly, thus resulting in a wrong estimate of the model’s true performance. One way to solve such problem potentially is to build better datasets with cleaner annotations. However, it is extremely hard and laborious to build a ``perfect'' dataset and many of the most widely used ones are all with different imperfections~\cite{haa500,Agrawal2018DontJA}. Especially, latest models are more and more often being trained with large-scale web-scraped data~\cite{radford2021learning,Goyal2021SelfsupervisedPO}, which inevitably contains various kind of noise.

\begin{figure}[t]
     \centering
     \begin{subfigure}[h]{0.55\textwidth}
         \centering
         \includegraphics[width=0.9\textwidth]{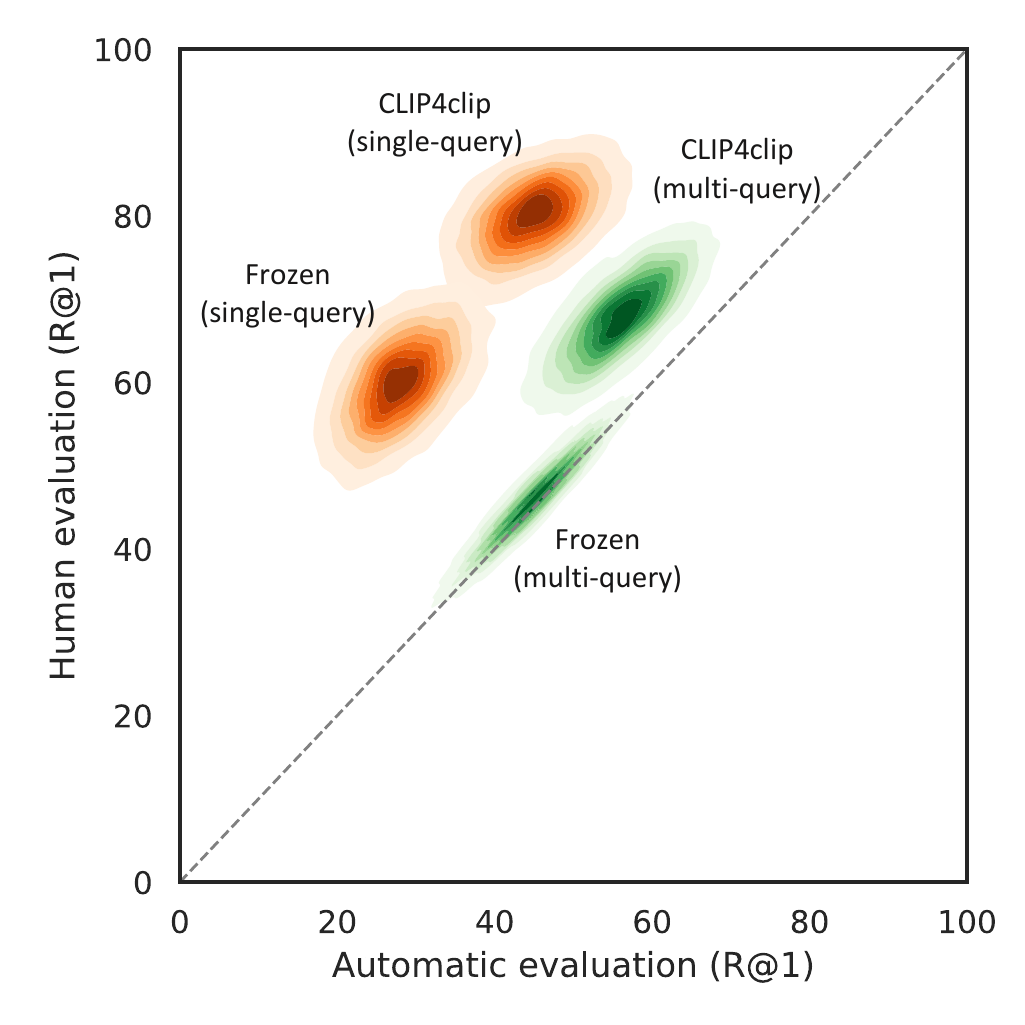}
         \label{fig:human_study}
     \end{subfigure}
     \hfill
     \begin{subfigure}[h]{0.42\textwidth}
         \centering
         \includegraphics[width=\textwidth]{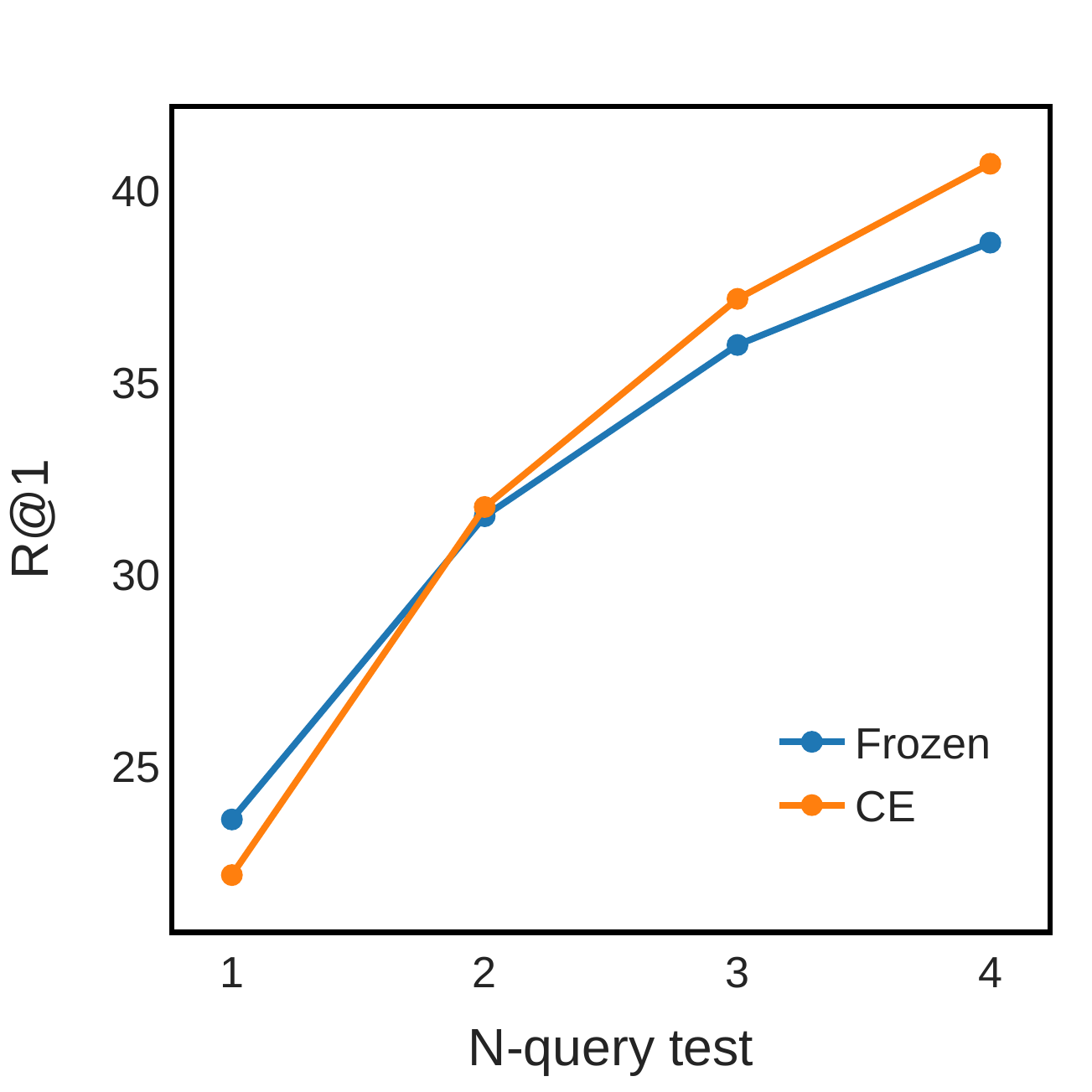}
         \label{fig:ce_vs_frozen}
     \end{subfigure}
     \vspace{-3mm}
    \caption{\textbf{Left}: Density plot of human evaluation vs. automatic evaluation for two video retrieval models~\cite{li2021clip,bain2021frozen} on MSR-VTT~\cite{xu2016msr}. Single-query retrieval is done by selecting the video with the highest similarity with the input query; for multi-query retrieval, the average similarity with all queries is used. Human evaluation (obtained by manual inspection of the top retrieved video) is compared to automatic evaluation (a binary indicator of whether the input queries are the corresponding annotations in the dataset for the top retrieved video). The results are computed on 100 retrieval samples, and the the plot is generated with Gaussian kernel density estimation on 10,000 samples generated with bootstrapping. Multi-query automatic evaluation shows significantly better correlation with human judgement compared to the single-query counterpart. \textbf{Right}: Relative performance of models~\cite{liu2019use,bain2021frozen} could change when evaluated with a different number of queries, suggesting that multi-query setting may reveal important insights about the retrieval potential of different models. The two models depicted are collaborative experts (CE)~\cite{liu2019use} and Frozen~\cite{bain2021frozen}; both models are trained from scratch on MSR-VTT~\cite{xu2016msr} for a fair comparison.
}
    \label{fig:single_vs_multi}
    
\end{figure}

    In this work, we propose the task of \textit{multi-query video retrieval} (MQVR) where multiple text queries are provided to the model simultaneously (Figure~\ref{fig:pull} bottom). It addresses the noise effects introduced by imperfect annotations in a simple and effective way. Intuitively, it is less likely that there are many videos in the dataset \emph{all} matching \emph{all} the provided queries. As a result, MQVR can produce a more accurate evaluation of the model's performance compared to a single-query retrieval benchmark. We verify this intuition empirically by inspecting 100 retrieval results, and manually labeling whether the top retrieved video matches the provided query or queries. As shown in  Figure~\ref{fig:single_vs_multi} left, single-query evaluation has a large gap with human judgement: automated methods unfairly penalize models for retrieving videos which actually match the provided query (as per human assessment) but are not the single video intended to be retrieved by the benchmark. Multi-query evaluation significantly reduces the gap by making the retrieval task less under-specified, so the model can better retrieve the target video (higher automatic evaluation score) and by making it less likely for other videos to match input queries. Going further, in Figure~\ref{fig:single_vs_multi} right we show that this has practical consequences as the relative ordering of two retrieval models changes under the single-query vs. multi-query evaluation settings.

    Equipped with these insights, we re-purpose existing video retrieval datasets and perform an extensive investigation of state-of-the-art models adapted towards handling multiple queries. We first experiment with different ways of using multiple queries only during inference, and then explore ways of learning retrieval models explicitly with multiple queries provided at training time. Our experiments over three different datasets demonstrate that MQVR more optimally utilizes the capabilities of modern retrieval systems, and can provide new insights for building better retrieval models. Especially, we find that a multi-query inspired training can lead to representations with superior performance and better generalization.

    To summarize, we make the following contributions:
    
    \begin{itemize}
        \item We extensively investigate multi-query video retrieval (MQVR) -- where multiple text queries are provided to the model -- over multiple video retrieval datasets (MSR-VTT~\cite{xu2016msr}, MSVD~\cite{chen2011collecting}, VATEX~\cite{wang2019vatex}). 
        \item We find that dedicated multi-query training methods can provide large gains (up to 21\% in R@1) over simply treating each query independently and combining their similarity scores. We also investigate several architectural changes such as localized and contextualized weighting mechanisms which further improve MQVR performance.
        \item To facilitate future research in this area, we further propose a new metric based on the area under the curve for MQVR with varying number of queries, which is complementary to standard metrics used in single-query retrieval.
        \item Finally, we also demonstrate that multi-query training methods can be utilized to benefit standard single-query retrieval, and that the multi-query trained representations have better generalization than the single-query trained counterpart.
    \end{itemize}

\section{Related work}
    \textbf{Video-text representation learning.} 
    The multi-modal learning for videos
    ~\cite{sun2019videobert,Arandjelovic2017LookLA,patrick2020support,wu2021exploring,yu2018joint,Hendricks_2017_ICCV,dong2021dual,wu2022snoc,gabeur2022masking,alayrac2020self} has been receiving increasing attention in recent years. 
    Among them, learning to generate better video and language representation
    ~\cite{zhang2018cross,gabeur2022masking,zhu2020actbert,wang2016learning} is one of the most attractive topics.
    Recent works leverage large-scale vision-language datasets for model pretraining~\cite{miech2018learning,bain2021frozen,li2020oscar,kamath2021mdetr}. Howto100M~\cite{miech2019howto100m} is a large-scale video-text pretraining dataset, which takes texts obtained from narrated videos using automatic speech recognition (ASR) as the supervision for video-text representation learning.
    Miech~\etal~\cite{miech2018learning} then proposed a multiple instance learning approach derived from noise contrastive estimation (MIL-NCE) to learn from these noisy instructional video-text pairs.
    Lei~\etal~\cite{lei2021less} proposed ClipBERT which leverages sparse sampling to train the model in an end-to-end approach.
    Recently, CLIP (Contrastive Language-Image Pre-training)~\cite{radford2021learning} has shown great success by pre-training on large-scale webly-supervised image and text pairs. The learned image and language representations are proved to be helpful in most downstream vision-language tasks.
    
    \textbf{Text-to-video retrieval.} To search videos by unconstrained text input, both video and text modalities should be embedded into a shared representation space for further similarity matching.
    Early works~\cite{mithun2018learning,miech2018learning,liu2019use,gabeur2020multi,wang2021t2vlad} used pre-trained models to extract representations from  multi-modal data (RGB, motion, audio).
    Mithun~\etal~\cite{mithun2018learning} proposed to use hard negative mining to improve the training for text and video joint modeling.
    Liu~\etal~\cite{liu2019use} leverage seven modalities (\eg speech content, scene texts, faces) to build the video representation. 
    Miech~\etal~\cite{miech2018learning} further proposed a strong joint embedding framework based on mixture-of-expert features.
    Gabeur~\etal~\cite{gabeur2020multi} introduce multi-modal transformer to jointly encode those different modalities with attentions. 
    Recently, there are some works~\cite{luo2021clip4clip,fang2021clip2video,li2021clip,gao2021clip2tv} based on the large-scale pretrained vision-language models (\eg CLIP~\cite{radford2021learning}), which achieve superior performance compared to previous training-from-scratch models.
    However, these works are evaluated using single-query benchmarks, suffering from the query quality issue.
    Differently, we study the multi-query video retrieval in this paper, where the input is a set of multiple queries.

    \textbf{Multi-query retrieval.} Many previous studies on multi-query retrieval are in the uni-modal domain, such as content-based image retrieval~\cite{vural2020deep,fernando2013mining,huang2017multi}, landmark retrieval~\cite{wang2017effective,wang2015effective}, or person retrieval using multiple images or video frames~\cite{zheng2016mars,liu2022unsupervised}. These methods take multiple images as input queries where each image is obtained from different angles, distances, conditions of the same objects or scenes~\cite{vural2020deep,wang2017effective}. Instead of getting multiple queries from the user, works on query expansion~\cite{chum2007total,chum2011total,imani2019deep,arandjelovic2012three,gordo2017end,radenovic2018fine,gordo2020attention} take a single query as input and further extend it to multiple queries by pre-processing (\eg synonym replacement) or using relevant candidates produced during an initial ranking. In this work, we utilize multi-query in the task of text-to-video retrieval to address challenges brought by imperfect annotations in the datasets.
    
\section{Multi-query video retrieval}

    In this section, we first formally define the setting of MQVR, then delineate the various methods experimented to handle multiple queries, and finally introduce a new evaluation metric for MQVR.

    \subsection{Setting}
    
    Video retrieval\footnote{Throughout this paper, we use the term `video retrieval' to refer the specific task of text-to-video retrieval.} is a task of searching videos given text descriptions~\cite{mithun2018learning,miech2018learning,liu2019use}. Formally, given a video database $\mathcal{V}$ composed of $n$ videos, $\mathcal{V} = \{v_1, v_2, ..., v_n\}$, and a text description $q_i$ of video $v_i$, the goal is to successfully retrieve $v_i$ from $\mathcal{V}$ based on $q_i$. This is the setting most existing video retrieval benchmarks adopt~\cite{xu2016msr,anne2017localizing,krishna2017dense,rohrbach2015dataset}. We refer to it as single-query setting, since one caption is used for retrieval during evaluation. Specifically, the goal is to learn a model $\mathcal{M}^{single}$ that can evaluate the similarity between any query-video pair, such that $\mathcal{M}^{single}(q, v) \in \mathbb{R}$ reflects how the query matches with the video. And a perfect retrieval would score the matching pair higher than non-matching pairs,
        \begin{equation}
            \forall j, j \neq i,~~~ \mathcal{M}^{single}(q_i, v_i) > \mathcal{M}^{single}(q_i, v_j).
        \end{equation}
        
    However, this single-query setting might be problematic when the input description has low quality, for example, being too general or abstract that can perfectly describe different videos in the database. This can pose serious trouble during evaluation. Due to the fact that existing video retrieval benchmarks are mostly based on datasets collected for other tasks (\eg video captioning)~\cite{xu2016msr,wang2019vatex,chen2011collecting,krishna2017dense,rohrbach2015dataset}, such low-quality queries are prevalent during model evaluation (as shown in Figure~\ref{fig:pull}). 

    To tackle this issue, we study the multi-query retrieval setting where more than one query is available during retrieval (Figure~\ref{fig:pull} bottom). Under similar notation, the task of multi-query retrieval is to retrieve a target video $v_i$ from the video database $\mathcal{V}$ based on multiple descriptions $Q_i = \{q_i^1, q_i^2, ..., q_i^k\}$ of it. And similarly we would like to learn a model $\mathcal{M}^{multi}$ that correctly retrieves the target video,
        \begin{equation}
            \forall j, j \neq i,~~~ \mathcal{M}^{multi}(Q_i, v_i) > \mathcal{M}^{multi}(Q_i, v_j).
        \end{equation}
    
    \subsection{Methods} 

        Broadly, we divide the methods into two categories: the first category of methods extends the models trained with single-query to multi-query evaluation in a post-hoc fashion without any retraining, and the second one has dedicated modifications for multi-query retrieval during training.
    
    \subsubsection{Post-hoc inference methods.}
    
        The models trained using single-query can be easily adapted when multiple queries are available, just by considering multiple queries separately and then aggregating the results.
        
        \paragraph{\textbf{Similarity aggregation (SA).}} A simple form is to take the mean of the similarity scores evaluated between each query and the video as the final multi-query similarity, which is then used to rank videos.
            \begin{equation}
                \mathcal{M}^{multi}_{SA}(Q, v) = \frac{1}{k} \sum_{i=1}^k \mathcal{M}^{single}(q_i, v).
            \end{equation}
        \paragraph{\textbf{Rank aggregation (RA).}} Unlike {\em similarity aggregation} which aggregates the raw similarity score of each query, {\em rank aggregation} aggregates the retrieval results instead. Denote $\mathcal{R}(v|q, \mathcal{M}^{single}) \in \{1,2, ..., n\}$ as the rank of $v$ among all videos in the candidate pool based on the evaluation of similarity to $q$ according to $\mathcal{M}^{single}$ (smaller rank means more compatible with the query), then the multi-query similarity can be calculated as:
            \begin{equation}
                \mathcal{M}^{multi}_{RA}(Q, v) = -\frac{1}{k} \sum_{i=1}^k \mathcal{R}(v | q, \mathcal{M}^{single}).
            \end{equation}
        Note that the overall similarity score here doesn't have a well-defined quantitative meaning as in the {\em \textbf{SA}} case, and just serves to order different videos.
        
    \subsubsection{Multi-query training methods.}
    
        Unlike post-hoc methods that essentially deal with the multi-query problem in a ``single-query way'', dedicated training modifications might be helpful if multiple queries are available during the training phase. This is already provided by many standard benchmarks~\cite{xu2016msr,chen2011collecting,wang2019vatex}, but to the best of the authors' knowledge, most existing works only treat the descriptions independently and adopt the single-query training, where a training text-video pair is composed of a video and one description, \ie ($q_i, v_i$) for a positive sample, ($q_i, v_j$) for a negative sample. However, as we will show later, this is not the best choice and that multi-query training, \ie using ($Q_i, v_i$) as a positive pair and ($Q_i, v_j$) as a negative pair, where $Q_i = \{q_i^1, q_i^2, ..., q_i^k\}$ contains more than one descriptions, can provide a large gain. 
    
        To facilitate the discussion, we first denote the similarity score of a query-video pair as $\mathcal{M}^{single}(q, v) = \mathbf{S}\left(f(q), g(v)\right)$. Here $f$ denotes the feature extraction network of query sentence which maps the input query to a high dimensional vector $f(q) \in \mathbb{R}^m$. Similarly, $g$, the video feature extractor, maps input video to the same embedding space $g(v) \in \mathbb{R}^m$. $\mathbf{S}$ is the metric that measures the similarity between $f(q)$ and $g(v)$ (\eg the cosine similarity). 
        
        \paragraph{\textbf{Mean feature (MF).}}  A naive way of combining multiple queries during training is to take the mean of their features to be the final query feature, with corresponding similarity score calculated as:
            \begin{equation}
                \mathcal{M}^{multi}_{MF}(Q, v) = \mathbf{S}\left(\frac{1}{k} \sum_{i=1}^k f(q_i),~g(v)\right).
            \end{equation}
        We will show in the next section that just by making this simple modification, the trained model can outperform the post-hoc methods by a large margin. However, a potential drawback of {\em mean feature} is that each query contributes equally to the result regardless of their quality. Thus we make a natural extension and further investigate \textbf{weighted feature (WF)} methods,
            \begin{equation}
                \mathcal{M}^{multi}_{WF}(Q, v) = \mathbf{S}\left(\sum_{i=1}^k \alpha_i f(q_i),~g(v)\right),
            \end{equation}
        \noindent where $\sum_{i=1}^k \alpha_i= 1$. We experiment with several ways to generate the weights $\alpha$.
        
        \paragraph{\textbf{Text-to-text similarity weighting (TS-WF).}} It is desirable for a query to contain complementary information about the target video that is not captured by other queries. On contrary, a query is not informative if it only contains redundant information already captured by others. This inspires a parameter-free method to evaluate the informativeness of one query by comparing the similarity of it to other queries. Specifically, the informativeness of query $q_i$ is computed as $\mathbf{I}(q_i) =  - \sum_{j \neq i} \mathbf{S}(f(q_i), f(q_j))$, where $q_i, q_j \in Q$. Notice the minus sign here means the informative queries should be different from others. Finally, the weights can be computed by taking softmax among $\mathbf{I}(q)$s.
        
        \paragraph{\textbf{Localized weight generation (LG-WF).}} The weights can also be learned using a separate network and trained end-to-end with other parameters. We experiment with a multi-layer perceptron (MLP) that maps extracted text features to a scalar. The MLP is shared across all queries and each query is processed separately. The resulting weights are normalized with softmax. 
            
        \paragraph{\textbf{Contextualized weight generation (CG-WF).}} Instead of computing weight for every query individually, {\em CG-WF} attends to other queries when generating each weight. Specifically, we experiment with a transformer-based attention network, where all query features are first input to a transformer to generate contextualized features, and then a MLP head is used to map the contextualized features to scalars. Softmax is used to compute the final normalized weights like in the {\em localized weight generation} case.

    \subsection{Evaluation} 
    
        We adopt the evaluation metrics widely used in the standard single-query retrieval setting and compute the standard R@K (recall at rank K, higher the better), median rank and mean rank (lower the better) in our experiments. Additionally, specific to multi-query evaluation, we would also like to compare models tested under varying number of query inputs (\eg Figure~\ref{fig:multi-query_test} in the experiments) for a more robust evaluation of the models' generalization ability. To facilitate future research in this subject, we  propose an area under the curve (AUC) metric. Specifically, $\mathbf{AUC}^{R@K}_{n}$ is defined as the normalized area under the curve value of R@K with test query-input varying from 1 to $n$,
        \begin{equation}
            \mathbf{AUC}^{R@K}_n = \frac{\mathbf{AUC}([R@K_1, R@K_2, ... R@K_n])}{n - 1},
        \end{equation}
    \noindent
    where $R@K_m$ is the R@K value when evaluating with m queries as input. 

\section{Experiments}

    We first describe in section~\ref{subsec:arch} the architecture backbones on which we build our multi-query retrieval experiment. Then specific experiment settings and implementation details are described in section~\ref{subsec:exp} and finally the experiment results are shown in section~\ref{subsec:res}.
    
    \subsection{Architecture backbones}
        \label{subsec:arch}
        \paragraph{\textbf{CLIP4Clip}~\cite{luo2021clip4clip}.} CLIP~\cite{radford2021learning} has been shown to learn transferable features to benefit lots of downstream tasks including video action recognition, OCR, etc. Recently, such trend has also been demonstrated for video retrieval. Among multiple models proposed in this line of work~\cite{andres2021straightforward,luo2021clip4clip,fang2021clip2video}, we adopt the CLIP4Clip for its simplicity. For our experiment, we use the publicly released ``ViT-B/32'' checkpoint for initialization of the CLIP model.
        \paragraph{\textbf{Frozen}~\cite{bain2021frozen}.} This model proposed by Bain \etal utilizes a transformer-based architecture~\cite{vaswani2017attention,dosovitskiy2020image}, which is composed of space-time self-attention blocks. For our experiments, we use the checkpoint provided by the original authors which is pretrained on Conceptual Captions~\cite{sharma2018conceptual} and WebVid-2M~\cite{bain2021frozen}.
        
    \subsection{Experimental setup}
        \label{subsec:exp}
        
                \paragraph{\textbf{Datasets.}}   We conduct our experiments on three datasets, which have multiple descriptions available for each video clip.
        
            \begin{itemize}[noitemsep]
                \item \textbf{MSR-VTT}~\cite{xu2016msr} is one of the most widely used datasets for video retrieval. It contains 10K video clips gathered from YouTube and each clip is annotated with 20 natural language descriptions. We follow the previous works to use 9K videos for training and 1K videos for testing.
                
                \item \textbf{MSVD}~\cite{chen2011collecting} is a dataset initially collected for translation and paraphrase evaluation. It contains 1970 videos, each with multiple descriptions in several different languages. We only use the English descriptions in our experiments. Following previous works, we use 1200 videos for training, 100 videos for validation and 670 videos for testing.
                
                \item \textbf{VATEX}~\cite{wang2019vatex} is a large-scale multilingual video description dataset. It contains 34991 videos, each annotated with ten English and ten Chinese captions. We only keep the English annotations and use the standard split with 25991 videos for training, 3000 for validation and 6000 for testing. 
                
            \end{itemize}

        \paragraph{\textbf{Implementation details.}} All the models are trained with 8-frame inputs and a batch size of 48 for 30 epochs. We adopt the cross entropy loss over similarity scores and a softmax temperature of 0.05 as used in~\cite{bain2021frozen}. AdamW~\cite{loshchilov2017decoupled} optimizer is used with a cosine learning rate scheduler and a linear warm up of 5 epochs~\cite{loshchilov2016sgdr}. The max learning rate is 3e-5 and 3e-6 for Frozen and CLIP4Clip respectively. We set query number to five for our multi-query experiments. Specifically, all available captions for each video are used during training, but for each training instance during forward pass, a subset of random five descriptions are input to the multi-query model. During test, we evaluate the N-query performance by sampling N query captions per video. However, to avoid selection bias, we repeat the evaluation for a hundred times (each time selects a different subset of N captions as queries for every video) and report the mean as the final results.
        
    \subsection{Results}
        \label{subsec:res}
        
                \begin{table}[t]
            \footnotesize
            
        \caption{Performance of different methods on MSR-VTT, MSVD and VATEX. The baseline is trained and evaluated with one query. Others are evaluated with five-query input. RA, SA are trained with one query. MF, TS-WF, LG-WF, and CG-WF are trained with five-query input. All numbers are the average over 100 evaluations with different query samples. Recall numbers are reported in percent.}
            \label{tab:main_res}

            \begin{subtable}[h]{\textwidth}
                \centering
                \begin{tabular}{lccccc|ccccc}
                  \toprule
                   &  \multicolumn{5}{c}{MSR-VTT~\cite{xu2016msr} (CLIP4Clip)} & \multicolumn{5}{c}{MSR-VTT (Frozen)}\\
                  \midrule
                    & R@1 $\uparrow$ & R@5 $\uparrow$ & R@10 $\uparrow$ & MdR $\downarrow$ & MnR $\downarrow$   & R@1 $\uparrow$ & R@5 $\uparrow$ & R@10 $\uparrow$ & MdR $\downarrow$ & MnR   $\downarrow$\\
                  \midrule
                  \underline{Baseline}         & 41.5 & 69.4 & 79.3 & 2.0 & 15.7
                                            & 31.9 & 61.1 & 72.6 & 3.0 & 23.2 \\

                  \midrule
                  RA              & 56.4  & 84.5  & 92.1  & 1.0  & 4.3
                                            & 44.6  & 76.2  & 85.2   & 2.0  & 7.0 \\

                  SA        & 68.4  & 92.1  & 97.0  & 1.0  & 2.4 
                                            & 55.3  & 85.2  & 92.5   & 1.0  & 4.2 \\

                  \midrule
                  MF              & 71.3  & 93.5  & 97.6  & 1.0  & 2.2 
                                            & 59.6  & 88.0  & 93.9  & 1.0  & 4.0 \\

                  TS-WF          & 72.6  & \textbf{94.2}  & \textbf{97.8}  & 1.0  & 2.1
                                            & \textbf{60.6}  & \textbf{88.5}  & \textbf{94.2}  & 1.0  & 3.8 \\

                  LG-WF     & 72.6  & 94.1  & 97.7  & 1.0  & 2.1
                                            & 59.6  & 87.6  & 93.8 & 1.0  & 3.9 \\

                  CG-WF     & \textbf{73.1}  & 94.1  & \textbf{97.8}  & 1.0  & 2.1
                                            & 59.9  & 88.1  & 94.0 & 1.0  & 3.7 \\    
                  \bottomrule
                \end{tabular}
                \label{tab:msrvtt_res}
            \end{subtable}
             
            \vspace*{0.5 cm}

            \begin{subtable}[h]{\textwidth}
                \centering
                \begin{tabular}{lccccc|ccccc}
                  \toprule
                   &  \multicolumn{5}{c}{MSVD~\cite{chen2011collecting} (CLIP4Clip)} & \multicolumn{5}{c}{VATEX~\cite{wang2019vatex} (CLIP4Clip)}\\
                  \midrule
                    & R@1 $\uparrow$ & R@5 $\uparrow$ & R@10 $\uparrow$ & MdR $\downarrow$ & MnR $\downarrow$   & R@1 $\uparrow$ & R@5 $\uparrow$ & R@10 $\uparrow$ & MdR $\downarrow$ & MnR   $\downarrow$\\
                  \midrule
                  \underline{Baseline}      & 43.8 & 74.0 & 83.0 & 2.0 & 10.9
                                            & 33.3 & 63.5 & 76.1 & 3.0 & 17.9 \\ 

                  \midrule
                  RA             & 47.3  & 79.9  & 89.8  & 2.0  & 4.8         
                                            & 43.6  & 75.5  & 86.1 & 2.0  & 7.9  \\ 
                                             
                  SA        & 57.6  & 87.1  & 93.7  & 1.0  & 3.4 
                                            & 47.3  & 77.9  & 87.9 & 2.0  & 6.7  \\
                  \midrule
                  MF             & 59.6  & 88.3  & 94.2  & 1.0  & 3.3     
                                            & 57.2  & 84.7  & 92.0 & 1.0  & 4.9  \\
                                             
                  TS-WF           & 60.4  & 88.8  & 94.5  & 1.0  & 3.3 
                                            & 58.4  & 85.3  & 92.5 & 1.0  & 4.6  \\
                                             
                  LG-WF     & 60.2  & 88.0  & 94.2  & 1.0  & 3.2    
                                            & 58.5  & 85.2  & 92.4 & 1.0  & 4.8 \\
                                             
                  CG-WF     & \textbf{61.7}  & \textbf{89.5}  & \textbf{94.9}  & 1.0  & 3.0  
                                            & \textbf{59.0}  & \textbf{85.6}  & \textbf{92.7} & 1.0  & 4.7  \\
                  \bottomrule
                \end{tabular}
                \label{tab:msvd_vatex_res}
            \end{subtable}

        \end{table} 
        
        Table~\ref{tab:main_res} summarizes the evaluation results on the test split of the MSR-VTT, MSVD and VATEX datasets. Due to space limit, we only show the {CLIP4Clip} numbers for MSVD and VATEX datasets. Please see the Appendix for the full results of {Frozen} model. Since the experiment findings are similar across both models, we will focus on the {CLIP4Clip} model during the following discussion.
        
        \paragraph{\textbf{(R1) Multi-query vs.\ single-query.}}
            The first row of Table \ref{tab:main_res} shows the result for the single-query baseline, where the model is both trained and evaluated with one query. The rest of the rows show the results for five-query evaluation. With five queries available, all models see a big boost in terms of retrieval accuracy compared to single-query. While it's anticipated that multi-query models should perform better than the single-query counterpart, such big improvement is still striking, and shows that single-query evaluation could underestimate the real retrieval abilities of models.
        
        \paragraph{\textbf{(R2) Post-hoc aggregation methods.}}
            Comparing two post-hoc methods which simply extend the pretrained single-query model to the multi-query setting, it's clear that {\em similarity aggregation} outperforms {\em rank aggregation} by a large margin. On {CLIP4Clip}, the R@1 is improved from 56.4\% to 68.4\% for MSR-VTT, from 47.3\% to 57.6\% for MSVD, and from 43.6\% to 47.3\% for VATEX, respectively. This is not surprising as the rank provided by a low-quality query is noisy and would thus drag down the final rank when combining with ranks provided by the other queries. While the similarity generated by a low-quality query is also noisy, its value tends to be small and is typically dominated by the similarity scores provided by the high-quality queries.

        \paragraph{\textbf{(R3) Multi-query training methods.}}
        
            It's clear from Table~\ref{tab:main_res} that multi-query training methods with ad-hoc training modifications improve a lot over the post-hoc methods. Just by simply feeding five captions and taking the mean of the encoded features as the final text features during training ({\em mean feature}), R@1 of {CLIP4Clip} model can be improved by 2.9\% (from 68.4\% to 71.3\%), 2.0\% (from 57.6\% to 59.6\%) and 9.9\% (from 47.3\% to 57.2\%) on MSR-VTT, MSVD and VATEX, respectively. We anticipate that such improvement can be mainly attributed to the de-noising effect of multi-query training. As the ranking loss acts on the combined features, the part of loss that tries to push apart the false-negative text-video pairs (due to general descriptions that match with more than one videos) is lessened, thus avoiding potential over-fitting and providing more robust features (additional evidence is also discussed in \textbf{R5}).
            
\begin{SCfigure}
        \centering
        \includegraphics[width=0.65\linewidth]{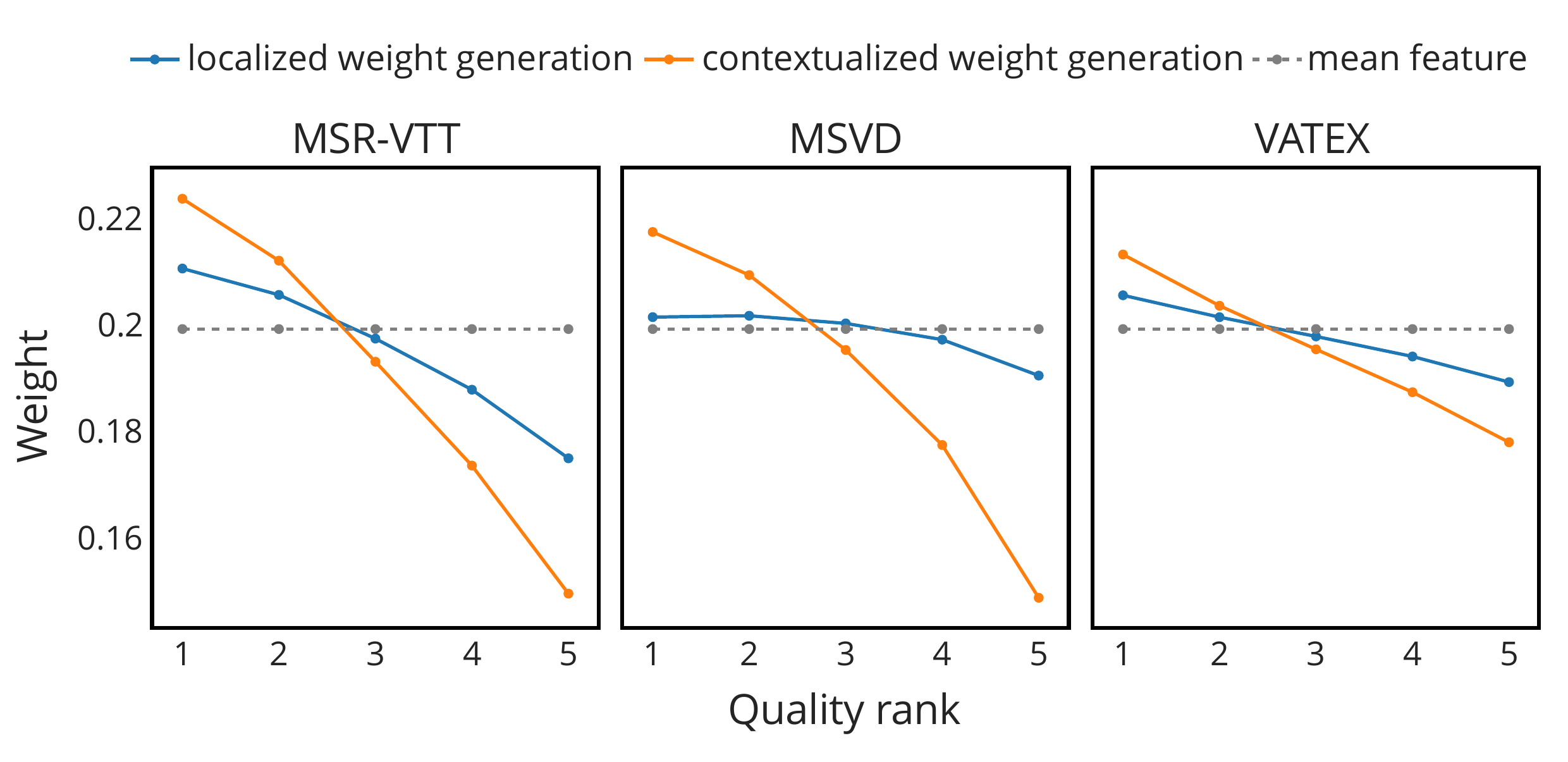}
        \caption{Averaged weights generated to five queries, which are ranked by their single-query retrieval results. 1 is the best and 5 is the worst. The average is taken over (test split size) * 100 instances.}
        \label{fig:weight_stats}
    \end{SCfigure}

\begin{figure}[t]
      \centering
      \includegraphics[width=1.0\linewidth]{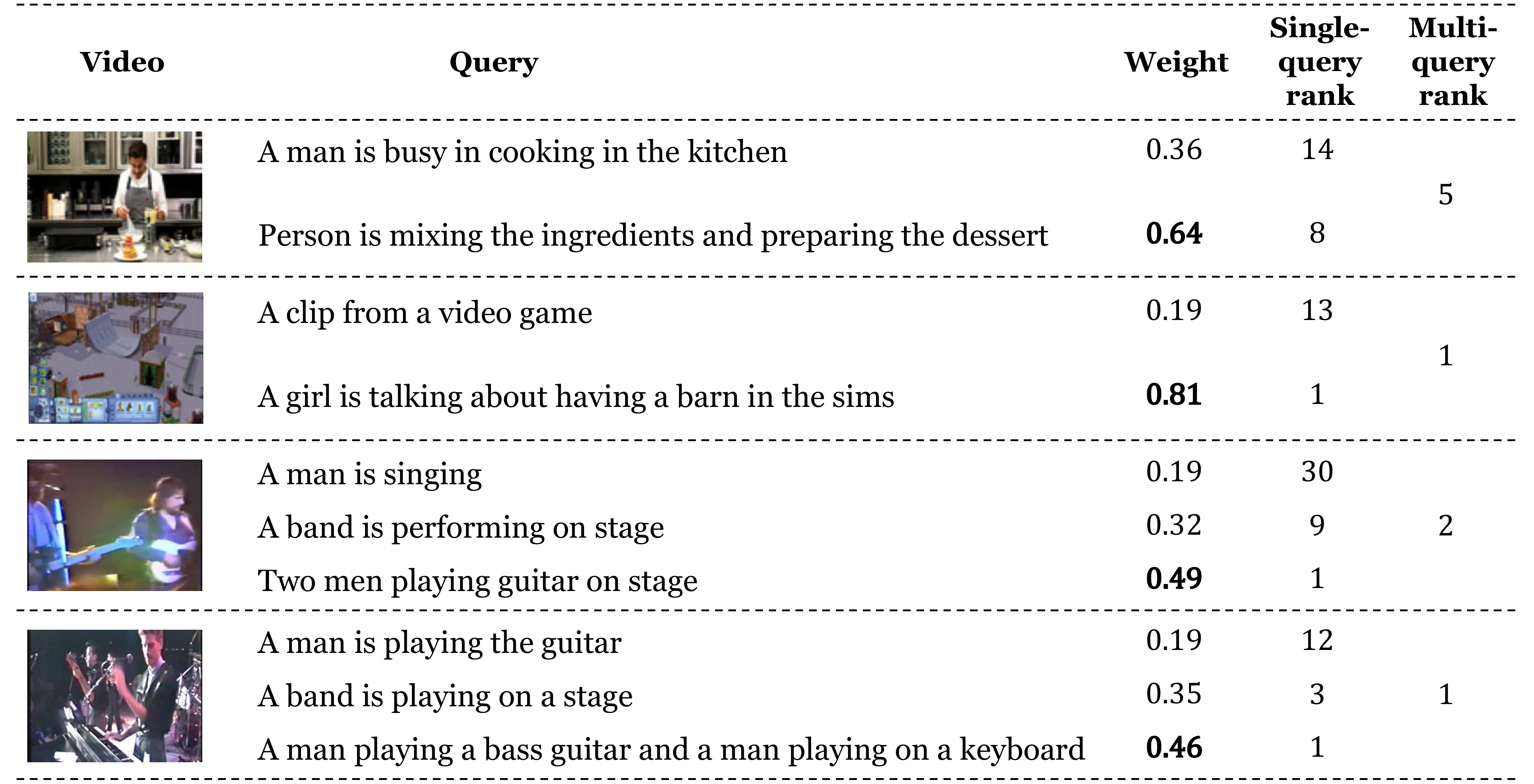}
      \caption{Qualitative examples of a weighted model (CG-WF) on MSR-VTT. It correctly attends to the disrciminative descriptions.}
      \label{fig:qualitative}
    \end{figure}

            Even though the {\em mean feature} training is already a very strong baseline, additional weighting heuristics still manage to introduce further improvements. On the {CLIP4Clip} model, with the best-performing {\em weighted feature} training with {\em contextualized weight generation}, R@1 can be improved over {\em mean feature} by 1.8\% (from 71.3\% to 73.1\%), 2.1\% (from 59.6\% to 61.7\%) and 1.8\% (from 57.2\% to 59.0\%) points across the datasets. To show that the learned weights can correctly capture the relative quality of the queries, we compute the average weights given to different queries ordered by their quality (we rank the quality of queries by their single-query retrieval result, 1 is the best, and 5 is the worst). Figure~\ref{fig:weight_stats} shows the result, and it's clear that the generated weights can correctly reflect the quality of the query. The {\em contextualized weight generation} works better than the {\em localized weight generation}. This is expected as instead of trying to learn a standalone quality predictor, the former lessens the task by learning to predict the relative quality. Especially when training data is small, \eg in MSVD, {\em localized weight generation} would have a hard time learning the quality and give more spread weights across different queries.  Figure~\ref{fig:qualitative} shows several qualitative examples on MSR-VTT for a model trained with {\em contextualized weight generation} (more examples shown in the Appendix). While some of the queries are of low-quality, the model can correctly attend to the better ones and achieve good overall retrieval accuracy.

        \paragraph{\textbf{(R4) Evaluation with varying number of queries.}}

\begin{figure}[t]
      \centering
       \includegraphics[width=0.95\linewidth]{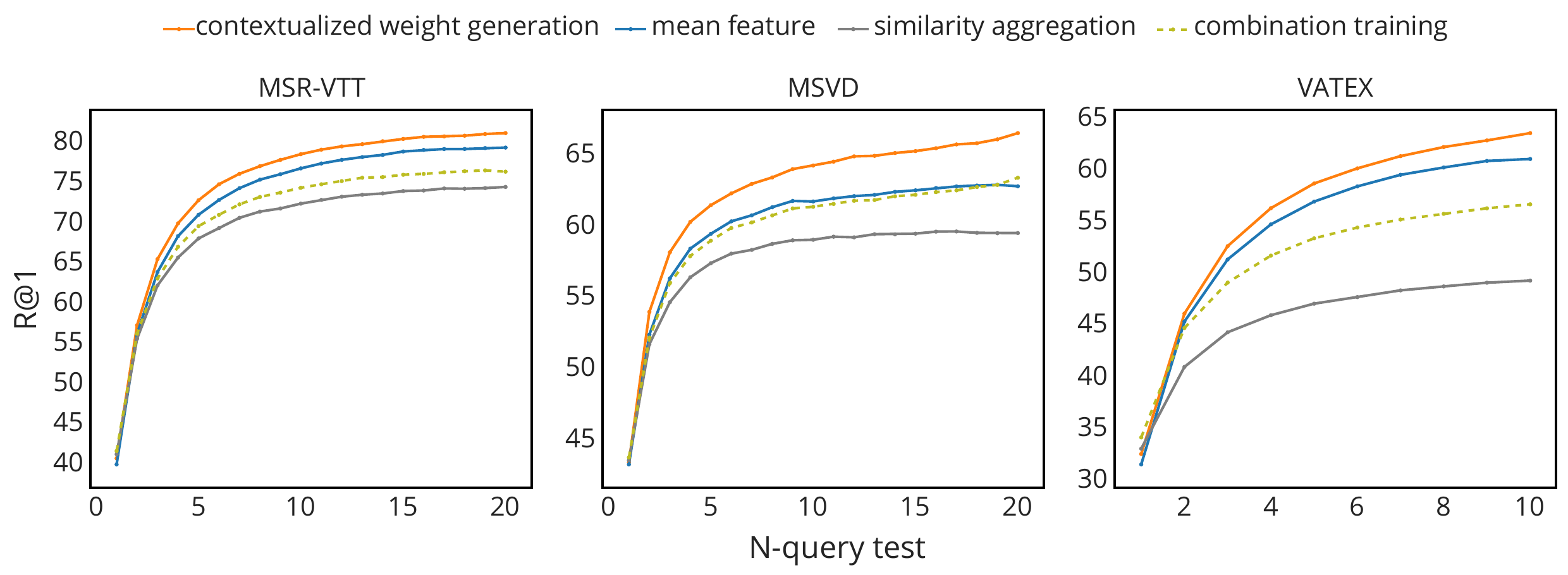}
       \caption{R@1 performance for SA, MF, CG-WF and {\em combination training} when evaluated with varying number of queries.}
       \label{fig:multi-query_test}
    \end{figure}

            Table~\ref{tab:main_res} shows the result where evaluation is performed under the same number of queries as multi-query models are trained (five-query training and five-query test). It is then natural to question whether such result can be generalized across varying number of queries. Figure~\ref{fig:multi-query_test} shows R@1 curves when the same models are tested under different number of queries and Table~\ref{tab:multi-query_test} summarizes the results in the form of the proposed $\mathbf{AUC}^{R@1}_n$ metric. First, the performance of all methods improves with more queries available. The curve increases rapidly at beginning and gradually saturates as more queries provides marginally additional information. Second, the five-query trained models outperform the single-query trained models across different number of testing queries except one, with dominating scores for $\mathbf{AUC}^{R@1}_3$, $\mathbf{AUC}^{R@1}_5$ and $\mathbf{AUC}^{R@1}_{10}$. This shows that multi-query training indeed learns better features suited for multi-query evaluation. While single-query trained models maintain the lead at single-query test, the differences are very small. Comparing {\em similarity aggregation} with {\em contextualized weight generalization}, the $R@1_1$ (single-query R@1) is 41.5\% vs.\ 41.0\% for MSR-TT, 43.8\% vs.\ 43.6\% for MSVD, and 33.3\% vs.\ 32.8\% for VATEX respectively. 
            
\begin{table}[t] 
    \centering
    \scriptsize
    \caption{Multi-query test results summarized with proposed $\mathbf{AUC}^{R@1}_n$ metric.}
    \label{tab:multi-query_test}
    
        \setlength{\tabcolsep}{1mm}
          \begin{tabular}{lccccc}
            \toprule
            & & $R@1_1$ & $\mathbf{AUC}^{R@1}_3$ & $\mathbf{AUC}^{R@1}_5$ & $\mathbf{AUC}^{R@1}_{10}$  \\
            \midrule
            
            \multirow{4}{3em}{MSR-VTT} & SA  & 41.5 & 54.0 & 59.9 & 66.1  \\
                                      & MF   & 40.3   & 54.4 & 61.3 & 68.8  \\
                                      & CG-WF  & 41.0  & \textbf{55.5} & \textbf{62.7} & \textbf{70.4}  \\\cmidrule{2-6}
                                      & Cmb (MF) & \textbf{41.9} & 54.6 & 60.8 & 67.5 \\
            \midrule
            
            \multirow{4}{3em}{MSVD}    & SA     & 43.8  & 50.6 & 53.5 & 56.4  \\
                                      & MF      & 43.5 & 51.3 & 54.8 & 58.3  \\
                                      & CG-WF   & 43.6 & \textbf{52.6} & \textbf{56.4} & \textbf{60.2}  \\\cmidrule{2-6}
                                      & Cmb (MF) & \textbf{44.0} & 51.2 & 54.5 & 57.9 \\
                            
            \midrule
            
            \multirow{4}{3em}{VATEX}   & SA     & 33.3  & 40.1 & 43.1 & 46.2 \\
                                      & MF      & 31.8 & 43.7 & 49.2 & 55.1 \\
                                      & CG-WF   & 32.8 & \textbf{44.6} & \textbf{50.4} & \textbf{56.8} \\\cmidrule{2-6}
                                      & Cmb (MF) & \textbf{34.4} & 43.5 & 47.6 & 52.1 \\
            
            \bottomrule
          \end{tabular}

\end{table}

            To get the best of both worlds, we further conduct an experiment with a combination of single-query and five-query {\em mean feature} training, \ie some training pairs contain one video and one caption while others contain one video and five captions (as with all other experiments, all available captions are used during training, but for each training instance, a random sample of one or five captions is used as input). The dash lines in Figure~\ref{fig:multi-query_test} show the results. Rather surprisingly, the {\em combination training} achieves the best single-query performance and outperforms single-query training entirely. This shows that \textbf{the de-noising effect of multi-query training can also be utilized to improve standard single-query retrieval}.
        
        \paragraph{\textbf{(R5) Training with different number of queries.}}

\begin{figure}[t]
     \centering
     \begin{subfigure}[b]{0.4\textwidth}
         \centering
         \includegraphics[width=\linewidth]{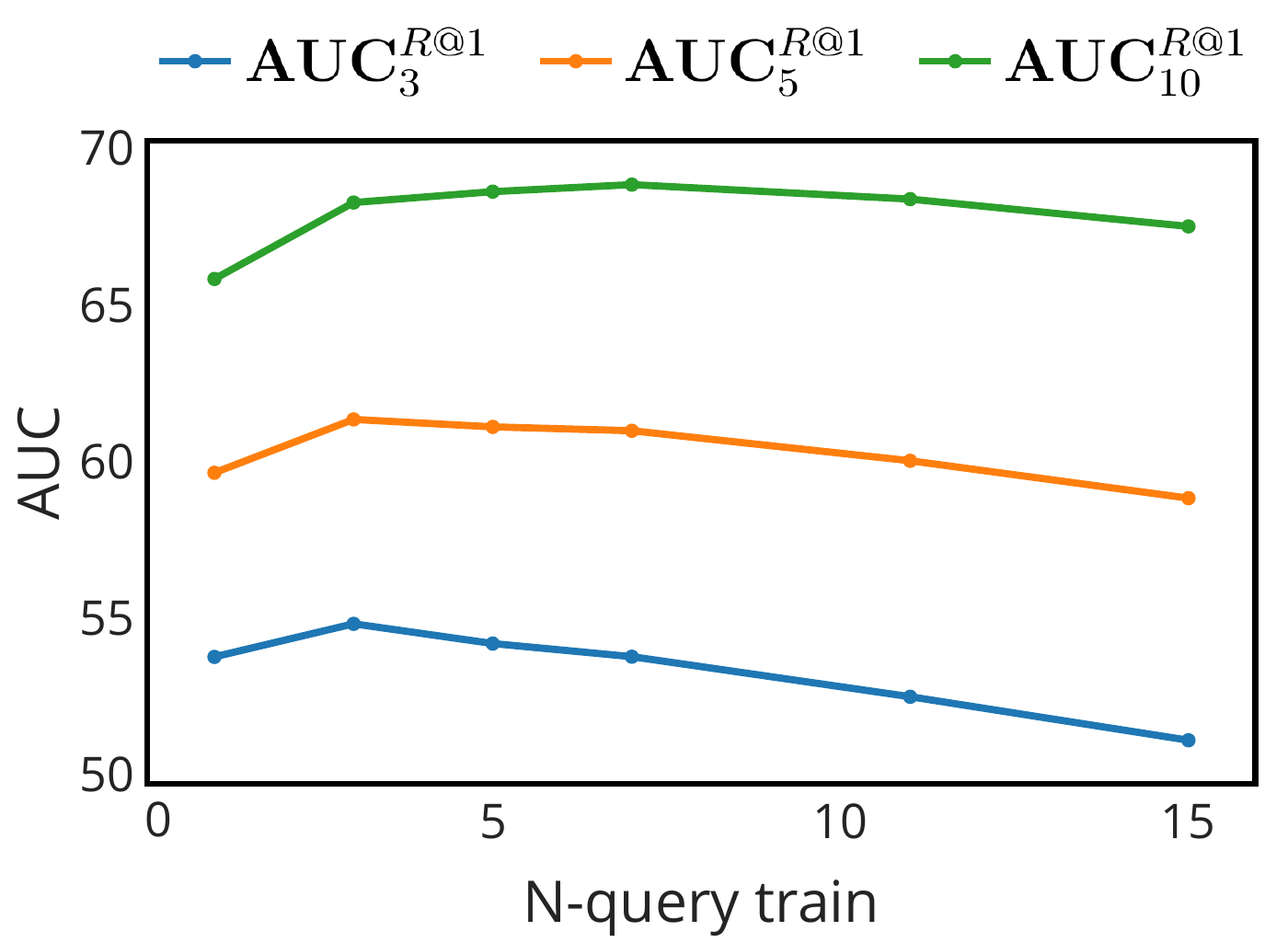}
           \caption{}
           \label{fig:different_query_training}
     \end{subfigure}
     \hfill
     \begin{subfigure}[b]{0.55\textwidth}
         \centering
         \scriptsize
         \begin{tabular}{lcccccc}
                \toprule
                
                Training    &  $R@1_1$ & $\mathbf{AUC}^{R@1}_3$ & $\mathbf{AUC}^{R@1}_5$ & $\mathbf{AUC}^{R@1}_{10}$ \\
                \midrule
                VATEX 1q   &       33.3 & 40.1 & 43.1 & 46.2 \\
                \midrule
                MSR-VTT 1q &       26.5 & 32.2 & 34.7 & 37.3 \\
                MSR-VTT 2q &       27.0 & 33.2 & 36.0 & 38.9 \\
                MSR-VTT 3q &       \textbf{27.1} & 33.8 & 36.9 & 40.0 \\
                MSR-VTT 5q &       \textbf{27.1} & 34.7 & 38.2 & 41.9 \\
                MSR-VTT 7q &       26.6 & \textbf{34.8} & \textbf{38.7} & \textbf{42.8} \\
                
                \bottomrule
              \end{tabular}
              \caption{}
             \label{tab:zero_shot}
     \end{subfigure}

    \caption{Evaluation of {\em mean feature} model trained on MSR-VTT with different number of queries. (a) In-domain test on MSR-VTT test set. (b) Out of domain zero-shot test on VATEX test set. ``$n$q'' means $n$-query training.}
    \label{fig:multi_q_train_and_zero_shot}
    
\end{figure}
            
            To understand the effect of number of queries (N) during training, we plot the performance of different models in Figure~\ref{fig:different_query_training}. We observe that the performance for all models increases at first, followed by a decrease (sometimes to a score even lower than N=1) as more queries are added. We hypothesize that this may be due to over-smoothing effect in the training process when too many queries are used, which causes the model to not learn discriminative representations for each individual query. 
            
            To further demonstrate the utility of the MQVR framework in learning robust representations, we perform a zero-shot transfer evaluation on VATEX dataset using models trained on MSR-VTT (Table~\ref{tab:zero_shot}). Quite promisingly, we observe that multi-query training (MSR-VTT 5q, MSR-VTT 7q) outperforms their single-query counterparts (MSR-VTT 1q) on both single-query and multi-query evaluations, achieving 0.6\% higher single-query R@1 and 2.6-5.5 points higher AUC, respectively. The latter results especially are only a few points below the AUC performance of an in-domain VATEX 1q model, demonstrating that there is a strong transfer and indicating that \textbf{multi-query training can lead to better generalization}.
            
\section{Conclusion}    

    In this work, we study the multi-query retrieval problem where multiple descriptions are available for retrieving target videos. We argue that this previously less-studied setting is of practical value both because it can provide significant improvement on retrieval accuracy through incorporating information from multiple queries and that it addresses challenges in model training and evaluation introduced by imperfect descriptions. We then investigate several multi-query training methods and propose a new evaluation metric dedicated for this setting. With extensive experiments, we demonstrate that multi-query inspired training methods can provide superior performance and better generalization.  We believe further investigation on the issue will benefit the field and bring new insights to building better retrieval systems for real-world applications.

\paragraph{\textbf{Acknowledgements.}} This material is based upon work supported by the National Science Foundation under Grant No. 2107048. Any opinions, findings, and conclusions or recommendations expressed in this material are those of the author(s) and do not necessarily reflect the views of the National Science Foundation. We would like to thank members of the Princeton Visual AI Lab (Jihoon Chung, Zhiwei Deng, William Yang and others) for their helpful comments and suggestions.

\clearpage
%
%
\bibliographystyle{splncs04}
\bibliography{main}

\clearpage

\appendix

\section{Additional experiment results}

\subsection{Five-query experiment for Frozen}

    Due to space limit, only MSR-VTT~\cite{xu2016msr} results are shown in Table~1 for Frozen~\cite{bain2021frozen} model. Here we include the results on MSVD~\cite{chen2011collecting} and VATEX~\cite{wang2019vatex} in Table~\ref{tab:frozen_res_msvd_vatex}. Similar findings can be drawn compared to the results on CLIP4Clip~\cite{luo2021clip4clip} model: the {\em similarity aggregation} outperforms {\em rank aggregation}. Dedicated multi-query training outperforms post-hoc inference methods. And that {\em weighted feature} training introduces additional improvement over {\em mean feature} training.

            \begin{table}[h]
            
            \caption{Performance of different multi-query retrieval methods on MSVD and VATEX datasets with Frozen~\cite{bain2021frozen} backbone. The baseline is trained and evaluated with one query. Others are evaluated with five-query input. RA, SA are trained with one query. MF, TS-WF, LG-WF, and CG-WF are trained with five-query input. All numbers are the average over 100 evaluations with different query samples. Recall numbers are reported in percent.}
                \label{tab:frozen_res_msvd_vatex}
            
                \centering
                \begin{tabular}{lccccc|ccccc}
                  \toprule
                   &  \multicolumn{5}{c}{MSVD~\cite{xu2016msr} (Frozen)} & \multicolumn{5}{c}{Vatex~\cite{wang2019vatex} (Frozen)}\\
                  \midrule
                    & R@1 $\uparrow$ & R@5 $\uparrow$ & R@10 $\uparrow$ & MdR $\downarrow$ & MnR $\downarrow$   & R@1 $\uparrow$ & R@5 $\uparrow$ & R@10 $\uparrow$ & MdR $\downarrow$ & MnR   $\downarrow$\\
                  \midrule
                  \underline{Baseline}         & 39.6  & 70.5  & 80.7 & 2.0  & 13.9
                                            & 26.7  & 56.9  & 70.4  & 4.0  & 26.4 \\

                  \midrule
                  RA              &  42.5  & 75.5  & 85.9  & 2.0  & 6.2
                                            &  37.1  & 69.8  & 82.1  & 2.0  & 11.7 \\

                  SA        & 53.2  & 85.3  & 93.0 & 1.0  & 3.7
                                            & 40.6  & 72.2  & 83.8  & 2.0  & 10.0 \\

                  \midrule
                  MF              &  55.8  & 86.5  & 93.8  & 1.0  & 3.7 
                                            &  49.0  & 79.2  & 88.3  & 2.0  & 8.1  \\

                  TS-WF          & 56.5  &\textbf{ 87.1}  & \textbf{93.9}  & 1.0  & 3.7
                                            & \textbf{49.7}  & \textbf{79.7}  & \textbf{88.7}  & 1.8  &     7.9 \\

                  LG-WF          & 56.4  & 86.3  & 93.4 & 1.0  & 3.9
                                             & 49.5  & 79.4  & 88.5  & 1.9  & 8.2 \\

                  CG-WF     & \textbf{56.7}  & 86.8  & 93.8  & 1.0  & 3.8
                                            & 49.4  & 79.3  & 88.5   & 1.9  & 8.6 \\    
                  \bottomrule
                \end{tabular}
                
            \end{table}

\subsection{Concatenation method for MQVR}

    Another straightforward method for handling multiple queries is to concatenate them together into a long sentence. The benefit of this approach compared to {\em mean feature} and {\em weighted feature} methods discussed in section 4 is that it can fully utilize the ability of (pretrained) language models, \ie information contained in multi-query is extracted directly by the language model, instead of using language model to extract information in each query separately and then combine them together. However, there are two downsides. First, this approach is limited by the maximum sequence length the language model can handle and concatenating multiple queries can quickly exceed that limit. For example, CLIP~\cite{radford2021learning} is pretrained with max sequence length of 77. So models built on it, like CLIP4clip~\cite{luo2021clip4clip}, can't handle longer sentence\footnote{We tried to extend the max length the model can handle by extending the pretrained positional embedding, but that leads to worse results.}. Second, as computation cost for transformer-based language model scales quadratically with increase of sequence length, {\em concatenation} method is less computation efficient.
    
\begin{figure}[h]
      \centering
       \includegraphics[width=1.0\linewidth]{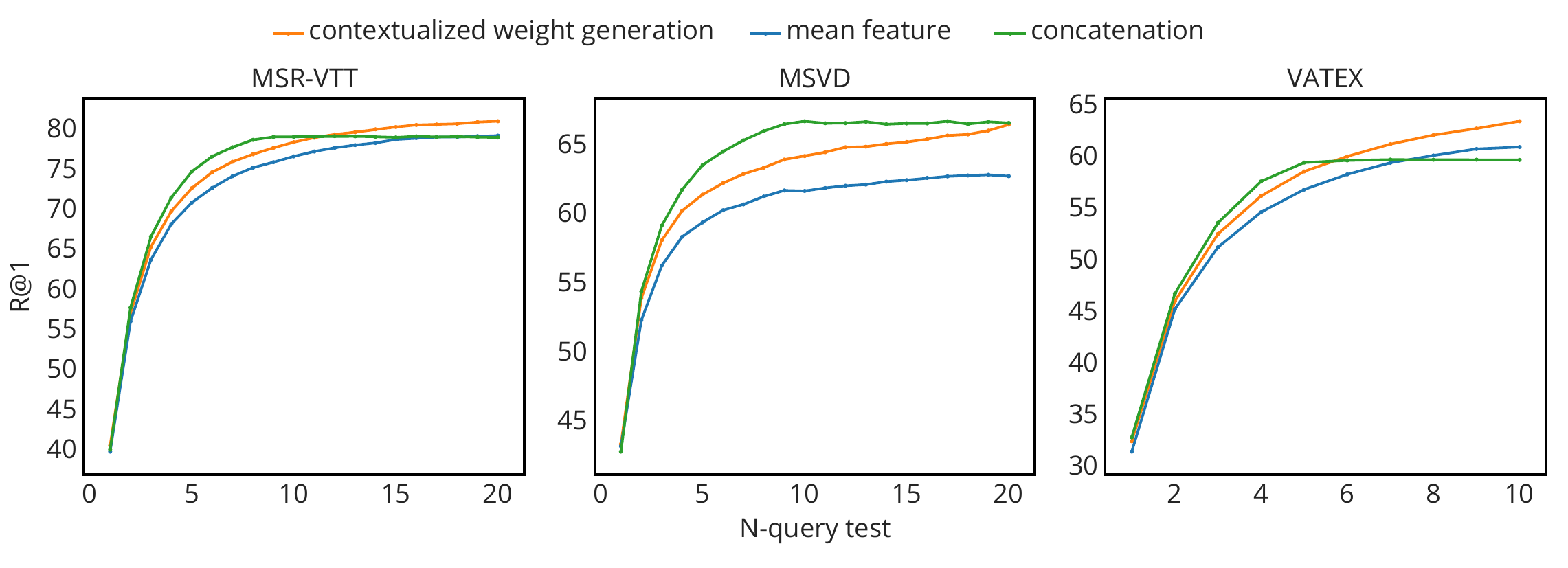}

       \caption{R@1 performance for {\em mean feature}, {\em contextualized weight generation}, and {\em concatenation}  when evaluated with varying number of queries.}
       \label{fig:concat_vary_test}
    \end{figure}

    Figure~\ref{fig:concat_vary_test} shows the comparison of {\em concatenation} with {\em mean feature} and {\em contextualized weight generation} for CLIP4clip~\cite{luo2021clip4clip} model trained with five-query~\cite{xu2016msr}\footnote{When the input sentence is longer than the max sequence length which can be processed by the language model, the sentence is cut off at max length and only the leading tokens are input to the model.}. As expected, {\em concatenation} outperforms the other two at low-query region. However, the performance stops increasing when more queries are available, at which point the concatenated sentence is longer than the max length the language model can handle.
    
\section{More qualitative examples}
    
\begin{figure}[t]
      \centering
      \includegraphics[width=1.0\linewidth]{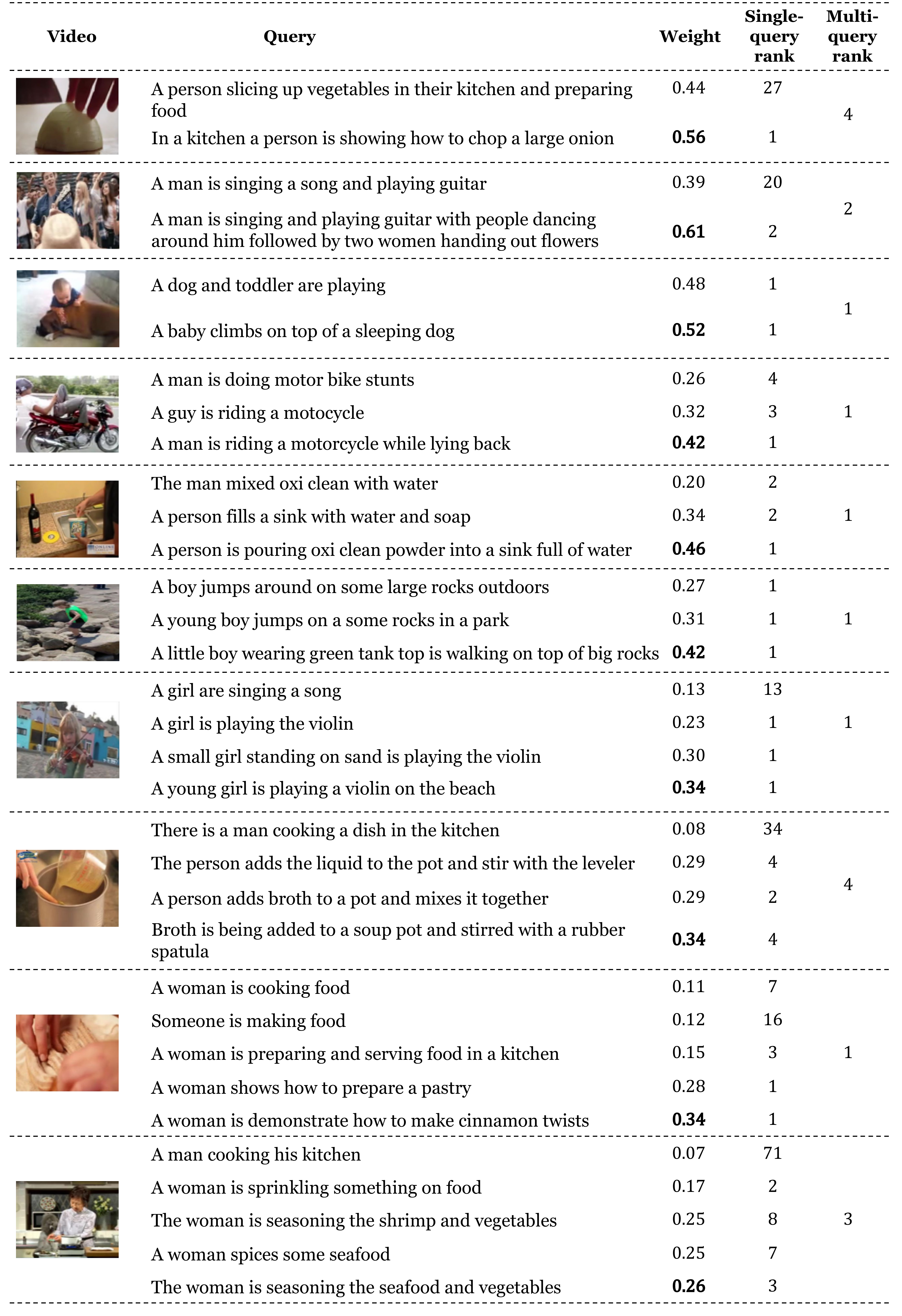}
      \caption{Qualitative examples of a {\em contextualized weight generation} model trained with five queries.}
      \label{fig:qualitative_appendix}
    \end{figure}

    Additional qualitative examples of generated weights for different queries are shown in Figure~\ref{fig:qualitative_appendix}. The weights correctly captures the relative quality among queries by giving higher weights to those containing more information.

\end{document}